\documentclass[a4paper]{article}
\usepackage{amsmath}
\usepackage{graphicx}
\usepackage{multirow}
\usepackage{url}
\usepackage{authblk}
\newcommand{\unit}[1]{\ensuremath{\, \mathrm{#1}}}
\newcommand{\specialcell}[2][c]{%
  \begin{tabular}[#1]{@{}c@{}}#2\end{tabular}}
\usepackage[english]{babel}
\usepackage[utf8x]{inputenc}
\usepackage{amsmath}
\usepackage{graphicx}
\usepackage[colorinlistoftodos]{todonotes}

\title{maxDNN: An Efficient Convolution Kernel for Deep Learning with Maxwell GPUs}
\author{
  Andrew Lavin
  \texttt{alavin@acm.org}
}
\affil{eBay Research Labs Machine Learning}
\begin{document}
\maketitle

\begin{abstract}
This paper describes maxDNN\footnote{The maxDNN source code can be downloaded from \url{https://github.com/eBay/maxDNN}}, a computationally efficient convolution kernel for deep learning with the NVIDIA Maxwell GPU. maxDNN reaches 96.3\% computational efficiency on typical deep learning network architectures. The design combines ideas from cuda-convnet2 with the Maxas SGEMM assembly code. We only address forward propagation (FPROP) operation of the network, but we believe that the same techniques used here will be effective for backward propagation (BPROP) as well.
\end{abstract}

\section{Introduction}

The central algorithm of convolutional neural networks is the 2D convolution of a bank of multi-channel filters against a minibatch of multi-channel 2D maps \cite{lecun2004learning}. Modern GPUs have demonstrated the ability to compute convolutions significantly faster than CPUs \cite{krizhevsky2012imagenet}. cuda-convnet2 and cuDNN are the leading GPU implementations of spatial domain convolution \cite{convnet-benchmarks}. fbcunn is a GPU implementation of frequency domain  convolution that has a speed advantage for many useful convolution shapes \cite{DBLP:journals/corr/VasilacheJMCPL14}.

When comparing convolution kernels, it is customary to report execution time or throughput. The problem with these measurements is that they conflate the separate issues of algorithm complexity, computational efficiency, and device peak throughput.

Computational efficiency is an interesting problem by itself, because it is in practice very difficult to write highly efficient GPU kernels. Anecdotal evidence suggests that for many real world problems such as SGEMM and convolution, it is not possible to create a GPU kernel with greater than 80\% computational efficiency using the CUDA Toolkit. The fact that the cuBLAS SGEMM kernel reaches 91\% computational efficiency on NVIDIA Maxwell GPUs using assembly optimization suggests that efficient kernels are possible on Maxwell, but that the CUDA Toolkit does not provide the necessary tools to create them.

The Maxas project is an open source assembler for NVIDIA Maxwell GPUs, created by Scott Gray \cite{maxas}. It includes as a programming example an assembly implementation of SGEMM that reaches 96\% computational efficiency.

We created maxDNN to demonstrate that the same techniques used by Maxas for generating efficient SGEMM machine code are also effective for convolution.

\section{Convolution}

The convolution used in deep learning takes a minibatch of $N_b$ 2D multi-channel maps of size $W_i \times H_i \times N_c$  and a bank of $N_o$ 2D multi-channel filters of size $S_k \times S_k \times N_c$. Convolution of $N_b$ maps against $N_o$ filters is performed separately in corresponding channels and the result is summed over all channels. A stride, possibly equal to 1, is chosen that yields the desired output map size, $W_o \times H_o \times N_o$. The origin of the convolution can be offset by an optional padding parameter, and the image is assumed to be wrapped in an apron of zeros to handle boundary spills. We also scale the results by a scalar, $\alpha$.

The algorithmic complexity of convolution is:

$$C(N_b, W_o, H_o, N_c, N_o, S_k) = 2 N_b W_o H_o (N_c S_k^2 + 1) N_o \unit{FLOPs}$$
where a single multiply-accumulate operation counts as 2 FLOPs.

\section{cuda-convnet2}

cuda-convnet2 \cite{cuda-convnet2} is perhaps the most efficient convolution across a wide variety of popular network shapes \cite{convnet-benchmarks}. It uses an interesting strategy for implementing direct convolution: the map and filter data is ordered in memory so that the inner dimension is the number of batches or filters. The outer dimensions are the width, height, and channels of the maps.

The calculation of $N_b \times N_o$ output values for a single output map coordinate is just a matrix multiply between an unrolled input map patch, of size $N_b \times S_k^2 N_c$, and the filter matrix, of size $S_k^2 N_c \times N_o$. 

Each of the $N_o$ columns of the filter matrix contains the $S_k^2 N_c$ weights for a single filter.

The $S_k^2 N_c$ columns of the input matrix must be gathered from non-contiguous segments of the input map. Column offset calculations used in the gather operation require extra integer instructions compared with the code used in general matrix multiply, which operates on contiguous blocks of matrix data.

cuda-convnet2 also employs effective GPU programming techniques, including:

\begin{itemize}
\item Textures to load global memory, reducing indexing arithmetic.
\item Loads and processes a tile of 8 columns of data per iteration, amortizing global load latency.
\end{itemize}

It is worth noting that cuda-convnet2 targeted the Kepler GPU, which has half (48K) the shared memory of the Maxwell GPU (96K). Extra shared memory makes possible additional latency hiding strategies.

\subsection{Maxas: The Maxwell Assembler}

Maxas is an open source assembler for the NVIDIA Maxwell architecture, created by Scott Gray \cite{maxas}. It gives the developer complete control over the scheduling of instructions and allocation of registers. The project includes an SGEMM implementation that reaches 96\% computational efficiency on Maxwell GM204 GPUs.

The project is interesting not just for the the assembler itself, but also for the SGEMM implementation and accompanying documentation, which is the best sample program published so far for creating high efficiency kernels. Several advanced techniques are used, including:

\begin{itemize}
\item Use of 128 bit texture load instructions to reduce number of global loads, to increase size of maximum array to 4GB (due to max texture index $2^{28}$), and to reduce indexing calculations
\item Double buffering of global memory loads to hide global memory latency.
\item Double buffering of shared memory loads to hide shared memory latency and reduce number of warp synchronizations per iteration
\item Coalesced storing to global memory by reorganizing output values through shared memory.
\item Zero initialization of registers using 128 bit shared memory loads.
\end{itemize}

The inner loop of Maxas SGEMM64 is 98.8\% floating point instructions (texture and shared memory load instructions are dual issued with arithmetic instructions and therefore are not counted). Occupancy is just 25\%, limited by the use of 127 registers per warp, so the high computational efficiency demonstrates the latency hiding power of instruction level parallelism.

The project contains SGEMM variants that perform $64 \times 64$ and $128 \times 128$ shared memory blocking. Both use $8 \times 8$ register blocking and load 8 columns of data per iteration.

\subsection{maxDNN}

The strategy behind our maxDNN kernel combines the style of cuda-convnet2 convolution with the matrix multiply assembly code of Maxas SGEMM. Maxas SGEMM64 was modified so that each block traverses a patch of the input map to compute a $64 \times 64$ filter-image block for a single output map coordinate. The z-coordinate of the block index is used to enumerate the filter-image blocks.

Basically this just required adjusting the indexing calculations in the existing SGEMM64 code. To reduce the number of indexing calculations required to traverse the input map patch, we lifted the calculation of pixel/channel offset locations into constant memory. The pixel offset is just added to the patch offset to compute the input map offset. This replaces the 3 nested loops over channels, rows, and columns of a patch with a single loop over all the precomputed offset locations. The result is an inner loop that is 98.3\% floating point instructions.

We also physically zero padded the input map to handle boundary overruns. We believe this restriction could be removed with a modest decrease in the percentage of floating point instructions.

\subsection{Experiments}

We compare the performance of maxDNN to cuDNN v.2 RC1 on a GEFORCE GTX980 graphics card which uses the NVIDIA Maxwell GM204 GPU. cuda-convnet2 was not used because it has not been optimized for the Maxwell architecture.

cuDNN v.2 RC2 was also available, but showed significantly worse performance, so we reverted to RC1.

We measure performance using computational efficiency, which is the ratio of the actual throughput of the program to the peak throughput of the device. The GM204 consists of 16 processors each with 128 cores. Each core is capable of executing 1 multiply-accumulate per clock cycle. So one can calculate the device peak throughput by

$$\text{Peak Throughput} = 2 \text{FLOPs}\cdot{128}\cdot{16}\cdot{\text{GPU Clock Speed}}$$

The factor of 2 is due to the custom of counting a single multiply accumnulate operation as 2 FLOPs.

Another way to measure computational efficiency is to divide the number of executed floating point instructions by 128, and then divide again by the number of processor clock cycles:

$$CE = \frac{1}{128} \frac{\text{fp instructions}}{{\text{processor clocks}}}$$
This appears to be the formula used by the flop\_sp\_efficiency metric in the nvprof profiling program in the CUDA Toolkit. It has the advantage of being independent of the clock speed, which can vary during kernel execution.

Using the above measure of efficiency, a kernel can get credit for unnecessary work by performing more floating point instructions than are strictly necessary. We see this arise as a modest effect in maxDNN when the filter size ($Sk^2N_c$) is not a multiple of the tile size (8). It has a more pronounced effect when the number of filters or mini-batch size is not a multiple of shared memory blocking size ($64\times64$).

Therefore we modify the computational efficiency function to only give credit for the number of FLOPs actually required by the direct convolution algorithm:

$$CE = \frac{1}{2\cdot{128}} \frac{C(N_b, W_o, H_o, N_c, N_o, S_k)}{{\text{processor clocks}}}$$

We report efficiency for two recent Imagenet contest winners, Alexnet (v.2) and Overfeat. The minibatch size for both networks is 128.

\subsection{Results}

Table \ref{table:alexnet_overfeat} compares the computational efficiency of cuDNN and maxDNN for FPROP convolution on the layers of Alexnet and Overfeat.

maxDNN efficiency for Alexnet v.2 ranges between 93.4\% and 95.5\%. The worst performance is on the input layer, where a patch only has $11\times11\times3$ elements. This reduces the size of the main loop, where almost all of the FLOPs are performed, compared with the initialization and storage code sections, which can be thought of as fixed overhead.

maxDNN efficiency for Overfeat reaches 96.3\%, and is over 94.4\% for all layers but the first, which scores just 70.3\%. This is due to the fact that the number of filters in this layer, 96, is not a multiple of the block size, $64\times64$. 

We believe this could be addressed with a kernel that uses a block size of $64\times32$. This would reduce the computational intensity with respect to global memory loads, but the high L2 cache hit rate of our kernel suggests there is a surplus of device memory bandwidth. Additional block sizes could be developed to accommodate small minibatch sizes. At a certain block size the computational intensity would be too low and the kernel would be device memory bandwidth limited, experiments are required to determine this threshold.

cuDNN efficiency on Alexnet varies between 32.5\% and 57.6\%. Not only are these numbers significantly lower, but the variance is much higher. For Overfeat the cuDNN efficiency varies from 39.6\% to 74.0\%. The first layer in each network is the least efficient, apparently due to a larger number of integer instructions used in indexing calculations.

One of the stated design goals of cuDNN was to achieve consistently high efficiency on a variety of convolution shapes using a single kernel \cite{DBLP:journals/corr/ChetlurWVCTCS14}. Although cuDNN reports flexibility with respect to minibatch size \cite{DBLP:journals/corr/ChetlurWVCTCS14}, we can see that in practice the performance varies a lot with respect to layer parameters.

\begin{table}
\begin{tabular}{ l l c c c r r r r }
Network & Layer & Input & Output & Kernel & \specialcell{cuDNN\\Efficiency} & \specialcell{maxDNN\\Efficiency} \\
\hline
\multirow{5}{*}{Alexnet v.2} & conv1 & 224x224x3 & 55x55x64 & 11x11 & 32.5\% & 93.4\% \\
 & conv2 & 27x27x64 & 27x27x192 & 5x5 & 46.7\% & 95.5\% \\
 & conv3 & 13x13x192 & 13x13x384 & 3x3 & 51.5\% & 95.1\% \\
 & conv4 & 13x13x384 & 13x13x256 & 3x3 & 57.6\% & 95.0\% \\
 & conv5 & 13x13x256 & 13x13x256 & 3x3 & 55.9\% & 94.6\% \\
\hline
\multirow{5}{*}{Overfeat} & L1 & 231x231x3 & 56x56x96 & 11x11 & 39.6\% & 70.3\% \\
 & L2 & 24x24x96 & 20x20x256 & 5x5 & 74.0\% & 95.6\% \\
 & L3 & 12x12x256 & 12x12x512 & 3x3 & 54.2\% & 94.4\% \\
 & L4 & 12x12x512 & 12x12x1024 & 3x3 & 62.9\% & 96.2\% \\
 & L5 & 12x12x1024 & 12x12x1024 & 3x3 & 63.2\% & 96.3\% \\
\end{tabular}
\caption{maxDNN convolution has consistently high efficiency for a variety of convolution shapes, providing the number of filters and minibatch size are both multiples of 64. cuDNN efficiency varies a lot between layers. These results are for FPROP operation with minibatch size 128.}
\label{table:alexnet_overfeat}
\end{table}

\begin{figure}[htbp]
\centering
\includegraphics[width=0.9\textwidth]{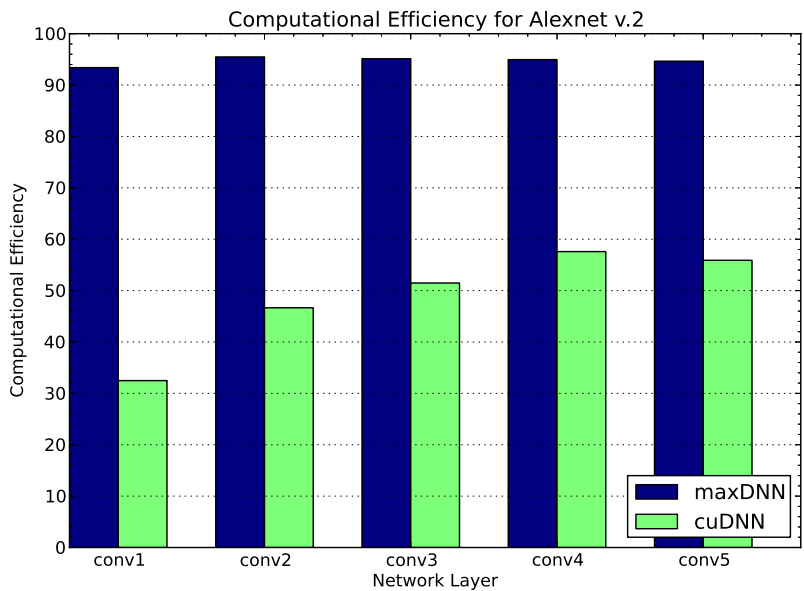}
\caption{FPROP convolution with minibatch size 128 for Alexnet v.2.}
\label{fig:alexnet}
\end{figure}
\begin{figure}[htbp]
\centering
\includegraphics[width=0.9\textwidth]{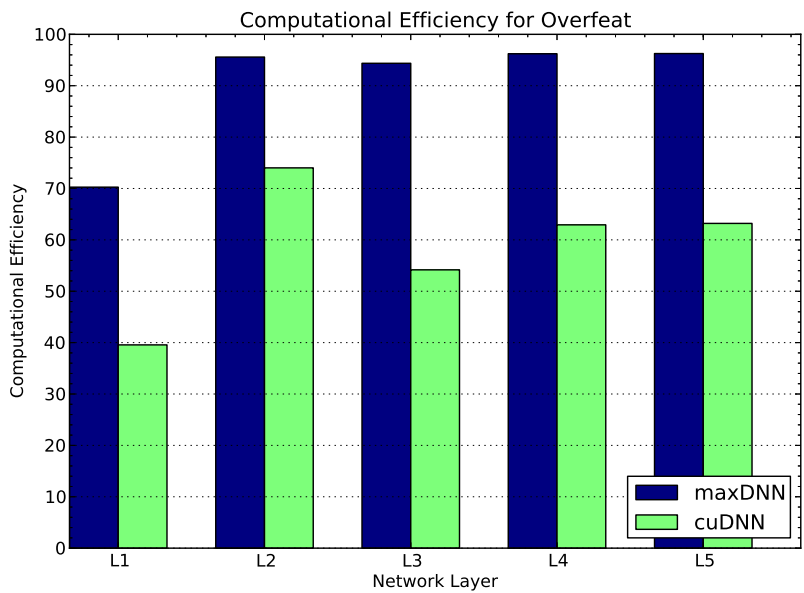}
\caption{FPROP convolution with minibatch size 128 for Overfeat. maxDNN efficiency suffers when the number of filters is not a multiple of 64, but is otherwise consistently high. maxDNN variants with other shared memory blocking sizes would likely address this shortcoming.}
\label{fig:overfeat}
\end{figure}

\subsection{Conclusion}

We developed an efficient convolution kernel for Maxwell GPUs using the Maxas assembler, Maxas SGEMM64 source code, and the cuda-convnet2 approach to convolution. We believe the same approach could be applied to the BPROP operation of convolutional neural networks.

The efficiency of maxDNN convolution rivals that of the best SGEMM implementations. Therefore maxDNN  represents an existence proof that high efficiency GPU convolution is possible.

\section*{Acknowledgement}

The author would like to thank eBay Research Labs Machine Learning Director Dennis DeCoste for his guidance during the course of this project.

\bibliographystyle{plain}
\bibliography{citations.bib}

\end{document}